# TOWARDS A THEORY OF GRANULAR SETS

Garimella Rama Murthy,
Professor,
International Institute of Information Technology,
Gachibowli, Hyderabad, INDIA

## 1. INTRODUCTION:

Set theory as a branch of human endeavour was developed by the efforts of many mathematicians [ Kam ]. Such a theory found many applications in science, technology and other fields. In an effort to capture uncertainity in human reasoning, Zadeh formulated and studied the theory of fuzzy sets. The theory of fuzzy sets found applications in many branches of science and technology. Pawlak, a computer scientist proposed and studied the concept of rough set in an effort to capture other aspects of uncertainty arising in applications such as database design. The theory of rough sets is found to have complemented the theory of fuzzy sets.

The author, in his research efforts related to the fusion problem in Wireless Sensor Networks ( WSN ) discovered the idea of "graded set" ( discussed in Section 2 ) as a generalization of the idea of rough set. When understanding the details of rough set theory, the author discovered the idea of "granular set". The basic motivation for such sets is discussed below.

Motivation for Granular Sets:

In biological systems such as trees, when the tissue is examined under a microscope at different resolutions, different cells/parts are observed. It is very clear that as the resolution increases, finer granular structure is observed. Our goal is to arrive at a mathematical abstraction of such sets observed in biological systems ( physical, chemical, biological etc ) as well as artificial systems ( such as databases ). It is expected that a detailed theory of such sets will find many applications as in the case of rough sets, fractal sets etc.

## 2. Wireless Sensor Fusion: Theory of Graded Sets:

- In the theory of rough sets, a set A is approximated by a tuple of sets ( $A_{lower}$ , $A_{Upper}$ ) where $A_{lower}$ is a subset of $A_{Upper}$ [4].

    Detailed theory of rough sets has been developed and applied in various fields of human endeavour. The author conceived the following generalization of a rough set called "graded set".

    **Definition of a Graded Set:**
    A = ( $A_1, A_2, ...., A_L$ ) where the $A_i$ sets satisfy the condition that they are a finite collection of nested sets i.e
    $A_i$ is a subset of $A_{i+1}$ for $1 \leq i \leq (L-1)$.

Such a set potentially arises in many applications. One such application is sensor fusion of interval valued measurements. Particularly we consider the "F" fusion function proposed by Schmidt et.al.

**Wireless Sensor Fusion: Graded Set:**

We now explain the approach to arrive at the fused interval based on finitely many measurement intervals.

Let $[a_1, b_1], [a_2, b_2], \ldots, [a_L, b_L]$ be the measurement intervals and let there be "f" faulty intervals ( we donot know which ones are faulty ). The fused interval estimate $[c, d]$ is obtained in the following manner [3].

c ( i.e. left end point of fused estimate ) is determined in the following manner. Arrange the left end points i.e. $a_i's$ from smallest to largest value. Counting down from the highest value i.e. $a_L$, consider the $f+1^{th}$ measurement. Consider it as 'c'.

d ( i.e. right end point of fused estimate ) is determined in the following manner. Arrange the right end points i.e. $b_i's$ from smallest to largest value. Counting up from the smallest value i.e. $b_1$, consider the $f+1^{th}$ measurement. Consider it as 'd'.

The fused interval estimate is [ c, d ]. It is clear that the end points c, d depend on the number of faulty intervals ( from say sensors ).

- We are interested in understanding how graded sets naturally arise when the number of faulty intervals ( from sensors ) is varied from a smallest to largest value.

Specifically, let $f \in \{f_{min}, f_{min}+1, f_{min}+2, \ldots, f_{max}\}$. From the above specified method of computing the left and right end points of fused estimate, we have the following result.

**Lemma 1:**

When f is increased from $f_{min}$ to $f_{max}$, the collection of fused intervals ( labeled as ) $A_i's$ lead to a graded set i.e.

$A_i$ is a subset of $A_{i+1}$ for $f_{min} \leq i \leq f_{max}$

i.e. they are nested intervals constituting a "graded' set.

**Proof:** The result follows from the definition of "F" fusion function of Schmidt et.al.

**Corollary:** The above result also applies to "M", "N" fusion functions i.e. When f is increased from $f_{min}$ to $f_{max}$, the fused estimates from "M", "N" functions constitute a graded set.

- Thus, in summary, with the number of faulty intervals/sensors as a parameter, the output of fused function is a graded set

- Suppose the number of faulty sensors assumes only two values i.e $\{f_{min}, f_{max}\}$, then the associated fused set constitutes a rough set.

*Random Graded Set*: *Definition:*

Consider the case where the number of faulty sensors is a discrete random variable. As a consequence, the output of "F"-fusion function assumes various intervals with associated probabilities. Hence when the number of faulty sensors is varied from the lowest to the highest value, we arrive at what we call as a RANDOM GRANULAR SET.

The above concept motivates the idea where the outcomes in a probability space are mapped to sets ( unlike the concept of a random variable where the outcomes are mapped to real/complex numbers ) i.e. the domain of mapping is outcomes in a finite/countable set and the range is sets.

## 3. Granular Sets: Information System:

In this section, we introduce the concept of "granular set" and discuss how such a set naturally arises in extraction of information from an information system.

Consider a finite set of objects / patterns i.e. $\{ x_1, x_2, \ldots, x_N \}$. Also, consider a collection of sets i.e. $C = [ A_1, A_2, \ldots, A_M ]$ each of which is a union of subsets of X. ( i.e. elements of $A_i \ for \ 1 \leq i \leq M$ are subsets of X ). Let $A_1$ be called the "finest set" and let $A_M$ be called the "coarsest set".

### Definition of a Granular Set :

With the above information, "C" is called a granular set if the elements of $A_{i+1}$ are either elements ( subsets of X ) of $A_i$ or a union of the elements of $A_i$ for all $1 \leq i \leq (M-1)$.

In other words, elements of $A_{j+1}$ are coarser granules than the elements of $A_j$ for $1 \leq j \leq (M-1)$.

Now we discuss how granular sets naturally arise in classification problems ( e.g. Pattern Recognition, databases ) associated with various applications.

We take an example information system with the associated information table. The table is taken from the WIKIPEDIA article on rough sets [2] :

SAMPLE INFORMATION SYSTEM

| OBJECT | $P_1$ | $P_2$ | $P_3$ | $P_4$ | $P_5$ |
|---|---|---|---|---|---|
| $O_1$ | 1 | 2 | 0 | 1 | 1 |
| $O_2$ | 1 | 2 | 0 | 1 | 1 |
| $O_3$ | 2 | 0 | 0 | 1 | 0 |
| $O_4$ | 0 | 0 | 1 | 2 | 1 |
| $O_5$ | 2 | 1 | 0 | 2 | 1 |
| $O_6$ | 0 | 0 | 1 | 2 | 2 |
| $O_7$ | 2 | 0 | 0 | 1 | 0 |
| $O_8$ | 0 | 1 | 2 | 2 | 1 |
| $O_9$ | 2 | 1 | 0 | 2 | 2 |
| $O_{10}$ | 2 | 0 | 0 | 1 | 0 |

**NOTE:** We now vary the set of attributes and determine the equivalence classes in an organized manner. Let the Target Set X be the set of all objects / patterns:

(I)     Let $P_E = \{ P_1, P_2, P_3, P_4, P_5 \}$.

Equivalence classes of the $P_E$-indiscernibility relation are denoted by $[x]_{P_E}$. They are given by

$C_5 = \{ \{ O_1, O_2 \}, \{ O_3, O_7, O_{10} \}, \{ O_4 \}, \{ O_5 \}, \{ O_6 \}, \{ O_8 \}, \{ O_9 \} \}$ .

(II)    Let $P_D = \{P_1, P_2, P_3, P_4\}$.

Equivalence classes of the $P_D$-indiscernibility relation are denoted by $[x]_{P_D}$. They are given by
$C_4 = \{\{O_1, O_2\}, \{O_3, O_7, O_{10}\}, \{O_4, O_6\}, \{O_5, O_9\}, \{O_8\}\}$.

(III)   Let $P_C = \{P_1, P_2, P_3\}$.

Equivalence classes of the $P_C$-indiscernibility relation are denoted by $[x]_{P_C}$. They are given by
$C_3 = \{\{O_1, O_2\}, \{O_3, O_7, O_{10}\}, \{O_4, O_6\}, \{O_5, O_9\}, \{O_8\}\}$.

(IV)    Let $P_B = \{P_1, P_2\}$.

Equivalence classes of the $P_B$-indiscernibility relation are denoted by $[x]_{P_B}$. They are given by
$C_2 = \{\{O_1, O_2\}, \{O_3, O_7, O_{10}\}, \{O_4, O_6\}, \{O_5, O_9\}, \{O_8\}\}$.

(I)     Let $P_A = \{P_1\}$.

Equivalence classes of the $P_A$-indiscernibility relation are denoted by $[x]_{P_A}$. They are given by
$C_1 = \{\{O_1, O_2\}, \{O_3, O_5, O_7, O_9, O_{10}\}, \{O_4, O_6, O_8\}\}$.

Now, we realize that the set $C = [C_1, C_2, C_3, C_4, C_5]$ is a *granular set* corresponding to the *graded set of features* $P = [P_A, P_B, P_C, P_D, P_E]$.

With the above example in mind, we arrive at the following Lemma.

**Lemma 2:** Let the target set be X, the collection of all objects / *patterns* and $Q = [Q_A, Q_B, ..., Q_L]$ be a graded set of features. Then the collection/ set of equivalence classes of the $\{P_A, P_B, ..., P_L\}$ indiscernibility relation constitute a granular set.

**Proof:** Follows from an interesting argument                          Q. E. D+

**Remark 1:** It should be noted that with X ( the set of all objects ) as the target set and "P" being chosen as one of the various possible graded sets, the associated collection of equivalence classes constitute a granular set.

**Remark 2**: The above discussion points to the fact that in any "hierarchical classification" effort ( as in the case of information table based classification in data bases ), the set that naturally arises is a "granular set".

- **Generalization of Rough Set Concepts: Graded Sets:**

  We now reason that, graded sets arise as a natural generalization of rough sets in classification problem.

  From the rough set theory, we have that the target set X can be approximated using only the information contained within P by constructing the P-lower ( denoted by $\underline{P}X$ ) and P-upper approximation ( denoted by $\overline{P}X$ ) of X:

  $\underline{P}X = \{ x \mid [x]_P \text{ is a subset of } X \}$

  $\overline{P} X = \{ x \mid [x]_P \cap X \neq \emptyset \}$.

  Now we are interested in understanding what happens if the target set, $X_i$ is varied in the following manner i.e.

  $X_i$ is a subset of $X_{i+1}$ for $1 \leq i \leq (L-1)$.

  i.e. $Y = [X_1 : X_2 : .... : X_L]$ is a graded set. From the above definitions of lower and upper approximations of P, it is clear that

  $\underline{P}\, X_i$ is a subset of $\underline{P}X_{i+1}$ for $1 \leq i \leq (L-1)$ and that

  $\overline{P}\, X_i$ is a subset of $\overline{P}\, X_{i+1}$ for $1 \leq i \leq (L-1)$.

  Thus, we necessarily have that the lower and upper approximations constitute "graded sets" when the target set constitutes a graded set.

- **Generalization of Information System:**

  We conceive of an information system which is represented by a three dimensional array ( and not a 2-dimensional matrix as given in the above example ). Thus the 3-dimensional array corresponds to "objects/patterns" versus "various sets of features". The feature sets could be dependent on one another. Thus, we have a collection of "information tables", that are all captured in a three-dimensional information array.

4. **Granular Sets : Applications:**

   As discussed previously, granular sets naturally arise in various hierarchical classification problems arising in various applications. Specifically, we realized that granular sets arise in classification problem associated with an information system represented by an information table. In this case, the sets arise when the target set is the whole set of objects and the collection of feature sets form a graded set.

   The granular sets obtained in this manner ( considering various possible feature sets that form a graded set ) provides a measure of sensitivity of classification of all the objects / patterns in X, as the set of features is incrementally increased. Since the target set can be a proper subset of X, the sensitivity of classification of objects / patterns with a variation of feature set can also be obtained. Some quantitative measures of sensitivity could easily be defined.

5. **Conclusions:**

   Motivated by the practical problem of wireless sensor fusion, the author introduced the concept of "graded set". Using graded set of features, it is shown that an interesting set called "granular set" naturally arises in classification problem associated with an information system. Using graded set of target sets, it is reasoned that the lower and upper approximations ( of the target set ) constitute a graded set of rough sets. Applications of "graded" as well as granular sets are actively being explored.